\documentclass{article}
\usepackage{ijcai22}

\usepackage{times}
\usepackage{soul}
\usepackage{url}
\usepackage[hidelinks]{hyperref}
\usepackage[utf8]{inputenc}
\usepackage[small]{caption}
\usepackage{graphicx}
\usepackage{amsmath}
\usepackage{amsthm}
\usepackage{amsfonts}
\usepackage{booktabs}
\usepackage{subcaption}
\usepackage{xcolor}
\title{Spatial-temporal Analysis for Automated Concrete Workability Estimation}

\author{
Litao Yu$^1$
\and
Jian Zhang$^1$
\and
Mohammed Bennamoun$^2$
\and
Xiaojun Chang$^1$
\and
Vute Sirivivatnanon$^1$
\and
Ali Nezhad$^3$
\affiliations
University of Technology Sydney$^1$ \\
The University of Western Australia$^2$ \\
Boral Australia Ltd.$^3$
\emails
\{litao.yu, jian.zhang, vute.sirivivatnanon\}@uts.edu.au$^1$ \\
mohammed.bennamoun@uwa.edu.au$^2$ \\
 ali.nezhad@boral.com.au$^3$
} 

\begin{document}

\maketitle

\begin{abstract}
Concrete workability measure is mostly determined based on subjective assessment of a certified assessor with visual inspections. The potential human error in measuring the workability and the resulting unnecessary adjustments for the workability is a major challenge faced by the construction industry, leading to significant costs, material waste and delay. In this paper, we try to apply computer vision techniques to observe the concrete mixing process and estimate the workability. Specifically, we collected the video data then built three different deep neural networks for spatial-temporal regression. The pilot study demonstrates a practical application with computer vision techniques to estimate the concrete workability during the mixing process.

\end{abstract}

\section{Introduction}

In Australia, approximately 28,000,000m$^3$ ready-mix concrete is delivered annually to construction sites by over 6,500 agitator (Agi) concrete-mixing trucks across the country. Physical infrastructure is critical to Australia’s social and economic function and comprises buildings and structures needed for the provision of transport, energy, water and communication. Concrete is predominantly used for the construction of Australia’s critical infrastructure and, therefore, its performance is vital to provide the nation’s essential services and maintain its economic activities. Workability is regarded as one of the most critical factors affecting the quality of finished concrete elements in buildings and infrastructure as well as the productivity and the costs associated with the construction of concrete structures. Inadequate workability negatively affects the ease of placement of concrete. This will commonly result in rejection of concrete deliveries at construction sites, leading to significant costs to concrete manufacturers and delay in construction projects. Concrete workability is controlled mainly by the content of water and admixtures which also simultaneously affect the mechanical and durability properties of concrete. Therefore, any effort to adjust the workability of concrete, if not implemented under adequate controls, may lead to undesired effects on concrete performance. In particular, unnecessary and uncontrolled increases in the water and/or admixture content of concrete, used to increase its workability, is translated into a decrease in concrete strength, which could compromise the performance of concrete buildings and infrastructure. The workability of concrete is quantified using a measure known as slump and is currently assessed through 1) visual inspection at the batch plant’s slump stand; and 2) the slump cone test on-site by a certified assessor. The visual inspection requires the assessor to ascend stairs in a wet heavy vehicle area to the top of the stand and inspect the mixing bowl through the back of an Agi truck, with a heavy flow of concrete pouring from the chute, which poses a real risk to worker’s safety and health. While lacking precision and being reliant on subjective interpretation of the assessor, this process is both labour expensive and prone to error.   

Computer vision techniques can provide an alternative approach to simulate the visual inspection for workability estimation to deliver multiple benefits including 1) reducing the risk of human error, 2) providing continuous real-time measurement of slump rather than discrete measurement at production and delivery points, 3) eliminating the health and safety risks. Contrary to visual inspection, computer vision techniques present a great opportunity to apply a computational model over time using video cameras, known as 3D Scanning, to monitor the concrete mixing. Despite some recent research on the application of computer vision for measuring the flow of concrete \cite{MDPI:KINECT}, there are no feasibility studies that apply computer vision and machine learning techniques for the concrete workability estimation in the mixing bowl. The challenge of modelling the above process lies in building a robust 3D Scanning (2D + T) of the concrete mixing based on the observed video data. 

In this work, we aim at a feasibility study on design and implementing a system for automated concrete workability estimation. The preliminary task involved annotating the captured video clips during the concrete mixing in a laboratory mixer and calibrating these images with manual slump cone \& flow tests \cite{ACI:SLUMP}. Through this task, we explore and identify the key techniques needed to develop the required capabilities, including visual and non-visual feature learning and multi-sensor fusion.

\section{Data collection and pre-procssing}

The core component of the vision-based model is data driven, which requires sufficient data for training and validation to obtain a reliable model in the automated system. To achieve this, we collected the video data as well as the annotation (workability measurement) in the Civil and Environmental Engineering Laboratory\footnote{\url{https://techlab.uts.edu.au/lab/civil-environmental-engineering-2/material-life-service/}} at UTS Tech Lab. To observe the concrete mixing process, we set up a mobile camera on a pole above the mixing bowl, which records the concrete mixing from the top-down view. 

\begin{figure}[t]\centering
  \begin{subfigure}{0.8\linewidth}
    \includegraphics[width=1\textwidth]{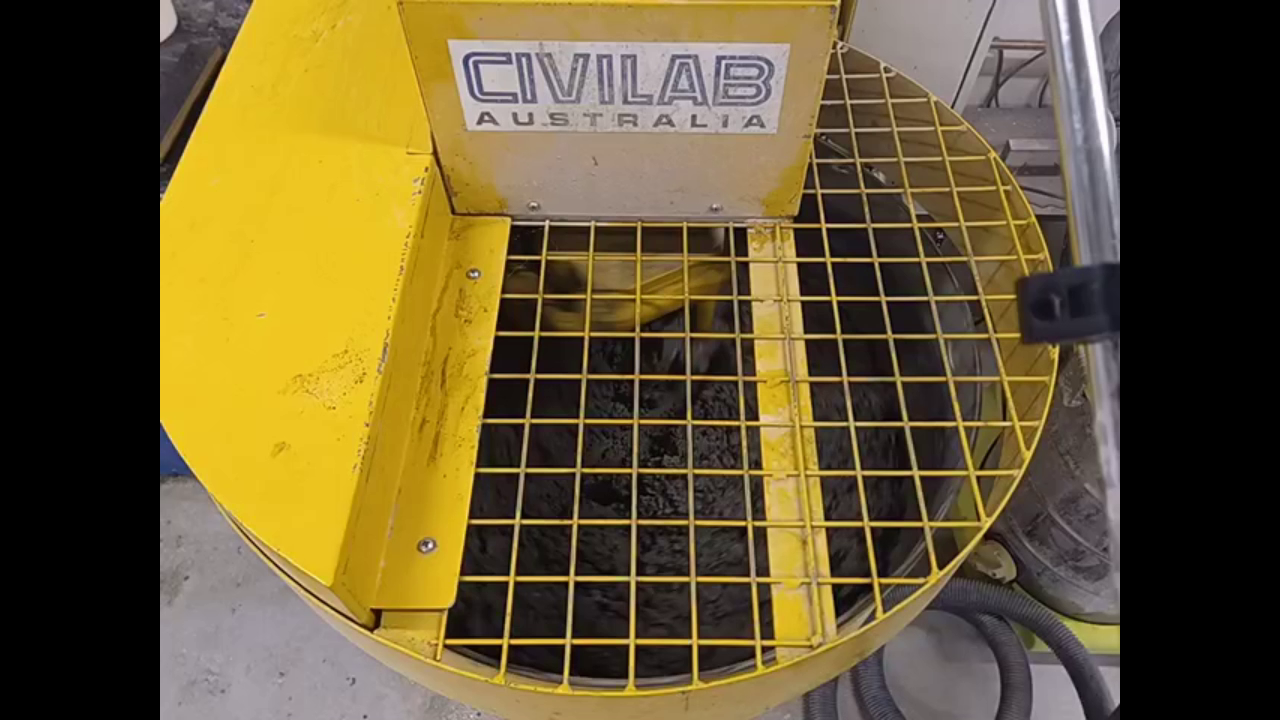}
    \caption{Mixing bowl.}
    \label{FIG1A}
  \end{subfigure}
  
  \begin{subfigure}{0.8\linewidth}
    \includegraphics[width=1\textwidth]{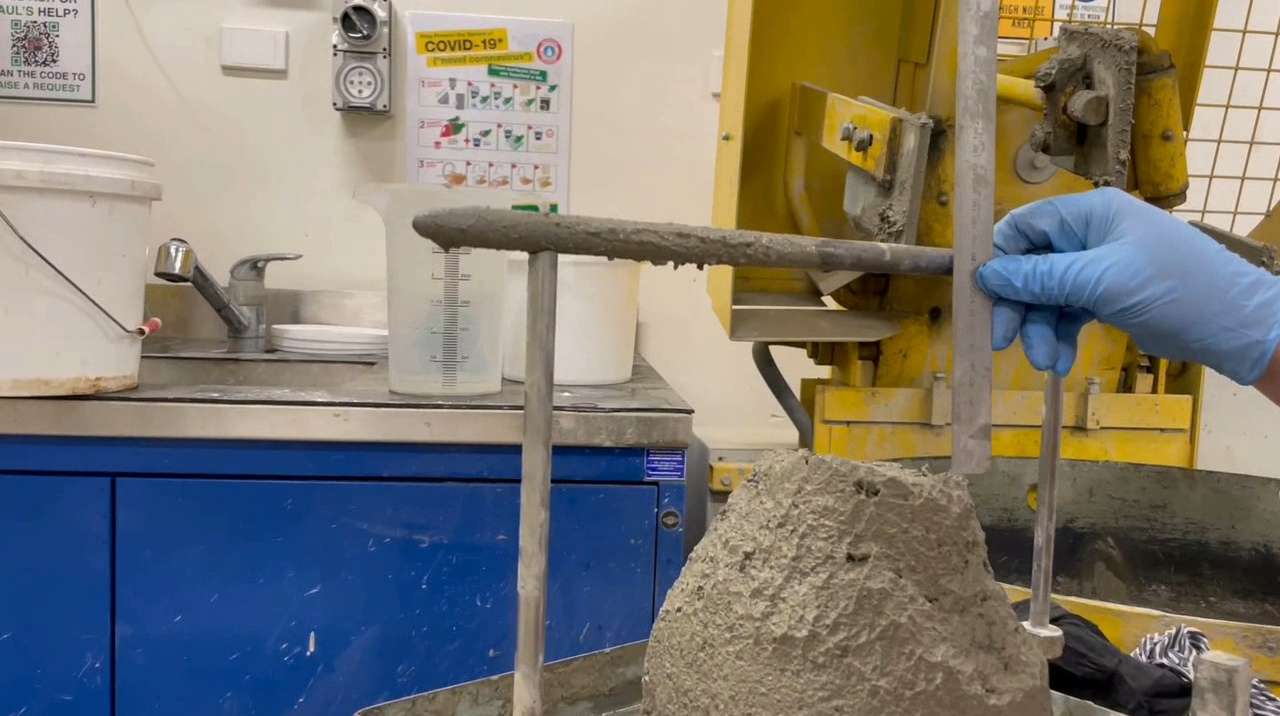}
    \caption{Slump-cone test.}
    \label{FIG1B}
  \end{subfigure}
  \caption{Video collection and ground-truth annotation.}
\end{figure}

We recorded each raw video and annotated the ground-truth in a mixing round. First of all, the mixing operator poured the cement, gravels, sand, water and other necessary indregients, into the mixing bowl. Then the bowl cap, equiped with a blender head, fell down and covered the top of the mixing bowl (see Figure \ref{FIG1A}). The video clip recorded the blending head working in the bowl until the concrete was evenly stired. Each round of mixing last about 2 minutes. After that, the operator sampled the concrete from the bowl and conducted the slump-cone test, using a ruler to measure the difference between the tops of slump and cone (see Figure \ref{FIG1B}). This height is used as the ground-truth to measure the workability, which usually reveals the stickiness of the fresh concrete. This procedure repeated by adding more water, or solid ingredients until the workability satisfies the requirement. In this feasibility study, we collected the videos from October to December in 2021, totally covering 9 complete mixing processes with 52 videos. 

We observed that for each video (about 2 minutes), the workability value is unstable during the blending because the cement, sand and other ingredients are not well mixed. This leads to the inaccurate annotation that negatively affects the model training. So for each video clip, we only used the final 10 seconds to ensure the data quality that corresponds to the slump height. Also, to ensure the observation area correctly focuses on the blending part, we applied the image segmentation on each frame, removing all other visual surroundings.

To prepare the data for model training, we sampled the frames from each video clip with 15 fps and a fixed 2 seconds' length. In the whole pilot study, we processed 255 video clips for the model training (185 videos), validation (35 videos) and testing (35 videos). Note that the current video dataset is far beyond training a robust and reliable model in real applications, due to the very limited number of data samples and the very strict visual environment. 

\section{Model building}

To build a real-time estimation system for concrete workability, we consider the following factors: 

\begin{itemize}

\item cleaned data representation;
\item fast inference; and
\item high-accuracy.
\end{itemize}

The system aims to simulate an assessor to conduct the visual inspection of the mixing process, which only focuses on the observable area in the bowl, because the surrounding visual information does not contribute to the workability estimation. So in order to provide the cleaned video data representation for model training and inference, we apply an image segmentation model to extract the key observation area, while the rest regions are all filtered by masking them to black colour. As such, it can effectively remove the noisy data and improve the model accuracy.   

We built deep neural networks for the concrete workability estimation. Since the workability measure is continuous, ranging from 40cm to 190cm, we consider it as a regression problem. Specifically, the model aims to learn to predict a value of slump height $\mathbf{y}$ given a short video clip, which can best approximate the slump-cone test $\hat{\mathbf{y}}$ measured by a ruler, i.e., $\min\|\mathbf{y}-\hat{\mathbf{y}}\|$. To achieve this, the model should be optimized to tune the parameters that simultaneously learn the visual feature representations and prediction functions in a stream-line.

Considering the very constrained visual surroundings within the key observation area, the regression model does not need to be over-parameterized that aims to learn very complex visual feature representations. Based on the collected video data with very low diversification, we design the following three models:

\begin{itemize}
\item {\bf Model-A} Time-distributed 2D convolution network,
\item {\bf Model-B} 3D convolution network, and
\item {\bf Model-C} 2D convolution LSTM network.
\end{itemize}

\begin{figure*}[t]
  \centering
    \includegraphics[width=0.8\textwidth]{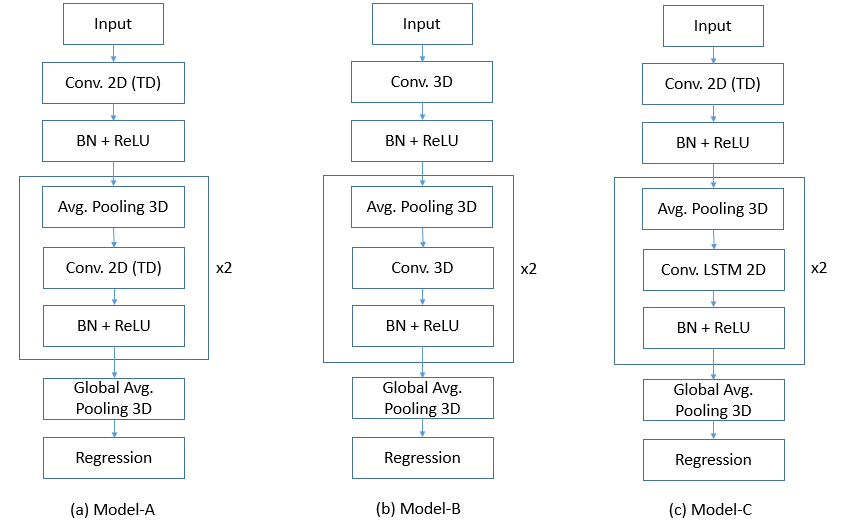}
    \caption{Network architectures.}
	\label{FIG:NETS}  
\end{figure*}

The learning architectures are illustrated in Figure \ref{FIG:NETS}. Here we test which model best fits the video data for acceptable accuracy with similar computational complexities. Model-A is a simple convolution network, without modeling temporal dependencies. Given an input video represented by a 3D tensor $t\times w\times h\times 3$, where $t,w$ and $h$ are the number of frames, width and height of the video clip, Model-A uses time-distributed 2D convolution for feature learning. In our experimental settings, $t=30,w=224,h=224$. Model-A consists of three convolution blocks, and each block has a time-distributed 2D convolution layer, a batch normalization layer and an activation layer (ReLU, rectified linear unit). The feature channels in the three convolutions are 16, 32 and 64, respectively. At the bottom of the architecture, we use a global average pooling layer and a dense-connected layer for regression. Model-B is also a convolution network, which replaces the time-distributed 2D convolution to 3D convolution. Such kind of learning structure can be used in action recognition \cite{TPAMI:3DCONV}. In Model-C, we use the LSTM (long-short term memory) to learn the temporal dependencies among the video frames. Specifically, we apply the 2D convolution LSTM \cite{NIPS:CONV_LSTM} in the 2nd and 3rd blocks. Although all of the three models are deep neural networks, they are practically very shallow with very limited trainable parameters (320K, 73K and 278K, respectively). Unlike the deep residual networks with more than 100 layers and much higher model complexities, pretrained on the very large dataset such as ImageNet \cite{CVPR:IMAGENET}, our models are directly trained on the collected video data.

The implementation is based on the fast-model building packages in TensorFlow v2.1. We employed the AdamW optimizer \cite{ADAMW} with the initial learning rate $10^{-4}$ in the training process, and the batch size was set to 16. Our experiments were conducted on a server equipped with a single NVIDIA Titan-X GPU card.

\section{Experimental results}

We carried out the experiment by training, validating and testing the three models and observed if they can well simulate assessors' expertise for concrete workability estimation. We used the Mean Absolute Error (MAE) as the objective function and evaluation metric, which describes the difference between the model inference and slump-cone test. The lower MAE is, the better performance the model can obtain.

\begin{figure}[t]
  \centering
    \includegraphics[width=0.4\textwidth]{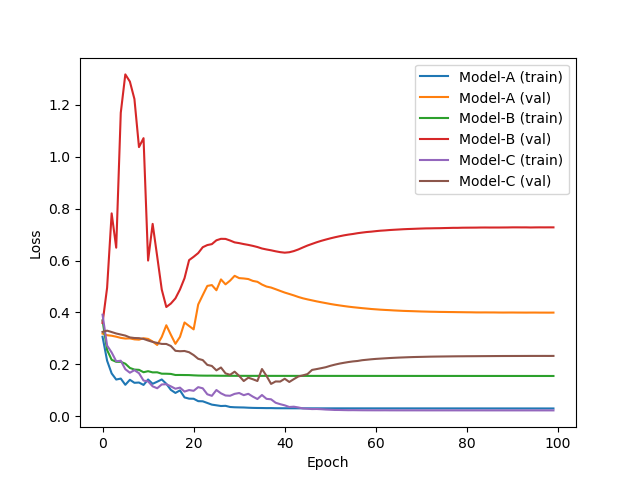}
    \caption{Convergence curves in training/validation.}
	\label{FIG:CONV}  
\end{figure}

We plotted the loss curves of the three models in Figure \ref{FIG:CONV}, which shows that all training loses converge quickly, because of the small-sized training dataset. However, when comes to the MAE on the validation dataset, neither Model-A nor Model-B can obtain a satisfactory estimation results. It is interesting that as a 3D convolution network that is able to learn the equal-sized temporal property, Model-B achieves the worse performance compared to Model-A, which only considers the static 2D information. Model-C performs the best among the three models, which proves the effectivity of combining 2D convolution and LSTM as the core function to learn spatial-temporal features for concrete workability estimation. At the same time, it is more parameter efficient that Model-A.  

\begin{table}[t]
\centering 
\caption{MAE results on 35 testing video clips.}
\label{TB:TEST}
\begin{tabular}{|c|c|c|}
\hline
{\bf Model-A}	& {\bf Model-B} & {\bf Model-C} \\
\hline
25.5cm $\pm$2.4cm & 31.3cm $\pm$2.6cm   & 10.8cm $\pm$3.3cm \\
\hline
\end{tabular}
\end{table}

We trained five rounds for each of the three models, with different parameter initializations. On the test split of the dataset, the experimental results are summarized in Table \ref{TB:TEST}. The three models obtain the MAE of 25.5cm, 31.3cm and 10.8cm, respectively. The standard deviations are similar, with 2.4cm, 2.6cm and 3.3cm, respectively. The preliminary results show that both the static appearance and temporal dependencies play important roles in accurately predicting workability. Although we have 135 video clips for training models, and the visual environment is highly constrained, the average 10cm prediction error is acceptable to demonstrate the effectiveness of the proposed method.

\section{Discussion and conclusion}

In this feasibility study, we have demonstrated a practical application of computer vision techniques to concrete workability estimation. The system dynamically monitors the mixing bowl and gives the workability estimation using a low-cost and easy-use setup. The three models used for this purpose are extremely simple, without the use of more curated methods, such as residual structure \cite{RESNET}, self-attention modules \cite{SE,HV,TRANSFORMER} and neural architecture search \cite{NAS}. These methods will be used when the collected dataset is large enough that contains much more diversified data samples. 

Also, the current practice of monitoring of concrete mixing process and assessing the concrete workability during the mixing should not only rely on the visual observation, but also some other signal channels such as concrete density, concrete weight, and hydraulic pressure. Based on these sensor data, we will also build the multi-feature fusion models \cite{MF}. With this regard, the automated system can well simulate the experienced assessor, to leverage each signal channel and alleviate the negative effects raised by the noisy data.

This research represents a technical breakthrough in our ability to monitor and assess the quality of the concrete product with minimal human interaction. Thus, the success of the forthcoming project is expected to significantly improve the performance of products in a large quantity, reduce waste and costs, and benefit Australian and global markets. 

\bibliographystyle{named}
\bibliography{ijcai22}

\end{document}